\documentclass[10pt]{article} 

\usepackage[preprint]{tmlr}

\usepackage{hyperref}
\usepackage{url}
\usepackage{bm}
\usepackage{bbm}
\usepackage{microtype}
\usepackage{graphicx}
\usepackage{subfigure}
\usepackage{booktabs} 
\usepackage{amsmath}
\usepackage{amssymb}
\usepackage{mathtools}
\usepackage{amsthm}
\usepackage{hyperref}
\usepackage[capitalize,noabbrev]{cleveref}
\usepackage[font=footnotesize]{caption}

\usepackage{siunitx}
\newcommand{\HCCnnTestAucSingle}{\num{0.922} (95\% CI: \num{0.917} to \num{0.926})}

\newcommand{\HCCreleaseNumFtr}{86}

\newcommand{\HCCtotPos}{49,380}
\newcommand{\HCCtotNeg}{1,094,011}

\newcommand{\HCCftrDecaySelNumFtr}{147}
\newcommand{\HCCftrDecayZeroNNAuc}{\num{0.926} (95\% CI: \num{0.921} to \num{0.930})}

\newcommand{\HCCftrDecayZeroNumFtr}{5610}

\newcommand{\PDACnnTestAucSingle}{\num{0.826} (95\% CI: \num{0.820} to \num{0.831})}

\newcommand{\PDACreleaseNumFtr}{83}

\newcommand{\PDACtotPos}{63,884}
\newcommand{\PDACtotNeg}{3,604,863}

\newcommand{\PDACftrDecaySelNumFtr}{121}
\newcommand{\PDACftrDecayZeroNNAuc}{\num{0.834} (95\% CI: \num{0.828} to \num{0.839})}

\newcommand{\PDACftrDecayZeroNumFtr}{5427}

\usepackage[textsize=tiny]{todonotes}

\newcommand{\eqnref}[1]{\hyperref[eqn:#1]{(\ref*{eqn:#1})}}

\newcommand{\real}{\mathbb{R}}
\newcommand{\V}[1]{\bm{#1}}
\newcommand{\vx}{\V{x}}
\newcommand{\vone}{\V{\mathbbm{1}}}
\newcommand{\nth}[1]{#1^{\text{th}}}
\newcommand{\attrf}[4]{#1(#2;\, #3,\, #4)}
\newcommand{\attr}[3]{\attrf{A}{#1}{#2}{#3}}
\newcommand{\attri}[4]{\attrf{A_{#1}}{#2}{#3}{#4}}
\newcommand{\defeq}{\mathrel{\raisebox{-0.3ex}{$\overset{\text{\tiny def}}{=}$}}}
\newcommand{\diff}{\mathop{}\!\mathrm{d}}
\newcommand{\condSet}[2]{\left\{#1 \;\middle|\; #2\right\}}

\newcommand{\se}{\mathfrak{X}}      
\newcommand{\mlsys}{\mathcal{M}}   

\theoremstyle{plain}
\newtheorem{theorem}{Theorem}[section]

\theoremstyle{definition}
\newtheorem{definition}[theorem]{Definition}

\theoremstyle{remark}

\title{Sound Explanation for Trustworthy Machine Learning\\
(Extended Abstract)}

\author{
    \name Kai Jia \email jiakai@mit.edu \\
      \addr Department of Electrical Engineering and Computer Science, \\
      Massachusetts Institute of Technology
      \AND
      \name Pasapol Saowakon \email zoomswk@mit.edu \\
      \addr Department of Electrical Engineering and Computer Science, \\
      Massachusetts Institute of Technology
      \AND
      \name Limor Appelbaum  \email{lappelb1@bidmc.harvard.edu} \\
      \addr Beth Israel Deaconess Medical Center
      \AND
      \name Martin Rinard \email rinard@csail.mit.edu \\
      \addr Department of Electrical Engineering and Computer Science, \\
      Massachusetts Institute of Technology
      }

\begin{document}

\maketitle

\begin{abstract}
    We take a formal approach to the explainability problem of machine learning
    systems. We argue against the practice of interpreting black-box models via
    attributing scores to input components due to inherently conflicting goals
    of attribution-based interpretation. We prove that no attribution algorithm
    satisfies specificity, additivity, completeness, and baseline invariance. We
    then formalize the concept, sound explanation, that has been informally
    adopted in prior work. A sound explanation entails providing sufficient
    information to causally explain the predictions made by a system. Finally,
    we present the application of feature selection as a sound explanation for
    cancer prediction models to cultivate trust among clinicians.
\end{abstract}

\section{Introduction}

As machine learning systems take more essential roles in decision making in
critical situations, there is increasing focus on understanding the reasoning
mechanism of the learned models \citep{molnar2020interpretable}. Prior work has
attempted to interpret black-box models in a post hoc manner \citep{
ribeiro2016should, zeiler2014visualizing, sundararajan2017axiomatic}, train
models that generate explanations besides predictions \citep{ park2018multimodal,
kumar2022self}, or build inherently interpretable models \citep{
andrews1995survey, sudjianto2021designing}.

However, there is no consensus on choosing the best interpretation algorithm if
the model is not inherently interpretable. The evaluation of post hoc or
self-explaining interpretation methods is often subjective \citep{
doshi2017towards}; there are systematic differences in how human individuals
approach decision making based on model interpretation \citep{
broniatowski2021psychological}. Different interpretation algorithms can generate
different results for practical applications, while practitioners do not seem to
have a principled way to resolve those differences \citep{
krishna2022disagreement}.

This paper argues that the above difficulties in designing and evaluating
interpretation algorithms partially arise from fundamentally conflicting
objectives of interpretability. In particular, we show that no attribution
algorithm for black-box models possesses all the following desirable properties:
specificity (unused inputs have a zero score), additivity (attribution scores
decompose over model addition), completeness (attribution completely explains
model output), and baseline invariance (rank of inputs is invariant to change of
baseline inputs).

Instead of pursuing other post hoc interpretation methods, we advocate an
alternative route towards model explainability. We formalize the concept of
\emph{sound explanation}. Briefly speaking, a sound explanation is a subset of
the information used by a machine learning system, perceived as understandable
by humans, that causally determines the system's output. A sound explanation is
always accurate and reliable in explaining how the system makes a prediction.
The challenge faced by system designers is how to design conceptually
understandable explanations. For example, a decision tree is a sound explanation;
however, a tree with thousands of variables and hundreds of levels is hardly a
satisfactory explanation to any human. The idea of sound explanation has been
informally adopted by prior work. \citet{ rudin2019stop} argues that black-box
models should be abandoned in favor of inherently interpretable models, which
naturally provide sound explanations. \citet{koh2020concept} proposes models
with concept bottlenecks, which is a type of sound explanation. Unlike prior
work, our research formally defines sound explanation instead of using heuristic
arguments.

We present a case in cancer prediction. We provide sound explanation via feature
selection on large datasets derived from Electronic Health Records (EHRs)
containing millions of patients and thousands of features. We design an
efficient algorithm for feature selection for neural networks. The final model
is then trained on the less than 100 selected features. We work with two cancer
types: Pancreatic Ductal Adenocarcinoma (PDAC, a common type of pancreatic
cancer) and Hepatocellular Carcinoma (HCC, a common type of liver cancer). The
selected features agree well with known PDAC and HCC risk factors, thus
providing confidence in our models among clinicians.

\section{Nonexistence of Ideal Interpretation}

This section considers a restricted form of interpretation that assigns an
importance score to each dimension of any given input relative to a fixed
baseline input. Such a score attribution formulation is closely related to cost
sharing that has been extensively studied in game theory. We follow the
notations of \citet{ friedman2004paths} and \citet{ sundararajan2017axiomatic}
to establish our results.

\begin{definition}[Interpretation]
    Let $F: \real^n \mapsto [0,\,1]$ be a continuously differentiable function,
    $\vx' \in \real^n$ be a baseline input, and $\vx \in \real^n$ be an
    arbitrary input. An interpretation is a function $A: (\real^n \mapsto [0,\,
    1]) \times \real^n \times \real^n \mapsto \real^n$ such that $\attr{\vx}{F}
    {\vx'} = (a_1,\, \ldots,\, a_n)$ with $a_i \in \real$ being the contribution
    of $x_i$ towards the value of $F(\vx) - F(\vx')$. For simplicity of
    notations, we define $\attri{i}{\vx}{F}{\vx'} \defeq a_i$. We call the value
    of $\attr{\vx}{F}{\vx'}$ an attribution of $F(\vx)$ to $\vx$ relative to a
    baseline input $\vx'$.
\end{definition}
Here $F$ is the function of interest, such as a deep neural network. A baseline
input is an uninformative input that typically satisfies $F(\vx')=0$, such as a
black image for image classification tasks, as suggested by \citet{
sundararajan2017axiomatic}. Given an input $\vx$, an interpretation assigns
importance scores to individual components of $\vx$ to explain the prediction
$F(\vx)$.

We now define the properties that an ideal interpretation should have.

\noindent\textbf{Specificity}
An interpretation should be specific to influential inputs: it should assign
zero scores to inputs that do not influence function values. An interpretation
without specificity may be adversarially manipulated by adding unused inputs.
Specificity is called ``dummy'' in \citet{friedman2004paths} and
``sensitivity(b)'' in \citet{ sundararajan2017axiomatic}.
\begin{definition}[Specificity]
    An interpretation $A$ satisfies specificity if for any function $F$ and
    integer $i\in[1, n]$ such that $\forall \vx\in\real^n,\,\delta\in\real:\:
    F(\vx) = F(\vx+ \delta \vone_i) $ where $\vone_i$ is all zero but one at the
    $\nth{i}$ dimension, then $\attri{i} {\vx}{F}{\vx'} = 0$.
\end{definition}

\noindent\textbf{Additivity}
An interpretation is additive if its attribution decomposes over function
addition. Additivity is a weak and natural constraint on the behavior of
interpretation.
\begin{definition}[Additivity]
     An interpretation $A$ satisfies additivity if for any $F_1$ and $F_2$,
     $\attr{\vx}{F_1+F_2}{\vx'} = \attr{\vx}{F_1}{\vx'} + \attr{\vx}{F_2}{\vx'}$.
\end{definition}

\noindent\textbf{Completeness}
An interpretation is complete if its attribution explains all of the output.
Completeness is also called efficiency in the cost sharing literature \citep{
friedman2004paths}.
\begin{definition}[Completeness]
    An interpretation $A$ satisfies completeness if $\sum_{i=1}^n \attri{i}{\vx}
    {F}{\vx'} = F(\vx) - F(\vx')$.
\end{definition}

\noindent\textbf{Baseline invariance}
The above properties have been studied by prior work. We propose a new property,
\emph{baseline invariance}. A baseline-invariant interpretation generates the
same rank of inputs for different choices of $\vx'$. Baseline invariance is
preferable because the widely used choice of $\vx'=\V{0}$ \citep{
sundararajan2017axiomatic} is arbitrary. For example, for image classification
tasks, it is equally sensible to set $\vx'=\V{0}$ as a black image, $\vx'=\V{1}$
as a white image, $\vx'=\V{0.5} $ as a gray image, $\vx'=E_{\vx \sim D}\vx$ as
the mean of training data, or $\vx' \sim N$ according to some noise distribution
$N$. An ideal interpretation should give similar attributions for different
choices of baseline inputs.
\begin{definition}[Baseline invariance]
    An interpretation $A$ satisfies baseline invariance if for any baseline
    inputs $\vx'_1 \in \real^n$ and $\vx'_2 \in \real^n$ and any function $F$
    such that $F(\vx'_1) = F(\vx'_2)$, it holds that for any indices $i$ and $j$,
    $\attri{i}{\vx}{F}{\vx'_1} < \attri{j}{\vx}{F}{\vx'_1}$ iff $\attri{i}{\vx}
    {F}{\vx'_2} < \attri{j}{\vx}{F}{\vx'_2}$.
\end{definition}

\begin{theorem}
    \label{thm:no-ideal}
    There is no interpretation $A$ satisfying all of these properties:
    specificity, additivity, completeness, and baseline invariance.
\end{theorem}

Our proof depends on the path method. The path method is a class of
interpretation methods that integrate the gradients of $F$ along a path. They
satisfy specificity, additivity, and completeness \citep{ friedman1999three}.
\begin{definition}[Path method]
    Given a continuous path function $\gamma: [0,\,1] \mapsto \real^n$ such that
    $\gamma(0) = \vx'$ and $\gamma(1) = \vx$, the path method
    $\attrf{A^{\gamma}}{\vx}{F}{\vx'}$ computes the attribution as
    \begin{align}
        \attrf{A^{\gamma}_i}{\vx}{F}{\vx'}
        \defeq \int_{t=0}^1 \partial_i F(\gamma(t)) \diff \gamma(t)
        \label{eqn:path-attr}
    \end{align}
\end{definition}

\begin{proof}[Proof of \cref{thm:no-ideal}]
    Assume $A$ satisfies specificity, additivity, and completeness. We prove
    that $A$ is inconsistent with baseline invariance. \citet[Theorem~1]{
    friedman2004paths} implies that given $\vx$ and $\vx'$, $A$ can be
    decomposed as a convex combination of path methods. Here it suffices to
    consider a single path method. Consider a linear function on $\real^2$:
    $F(\vx) = x_1 - x_2$, two baseline inputs $\vx'_1 = (-1,\, -1)$ and $\vx'_2
    = (1,\, 1)$, and the input $\vx_t = (1,\, 0)$. For two path functions
    $\gamma_1$ and $\gamma_2$ such that $\gamma_1(0) = \vx'_1$, $\gamma_2(0) =
    \vx'_2$, and $\gamma_1(1) = \gamma_2(1) = \vx_t$, their attributions are
    $\attrf{A^{\gamma_1}}{\vx_t}{F} {\vx'_1} = (2,\, -1)$ and
    $\attrf{A^{\gamma_2}}{\vx_t}{F}{\vx'_2} = (0, \, 1)$ (following
    \eqnref{path-attr}), which violates baseline invariance as the ranks of
    $x_{t1}$ and $x_{t2}$ in the two attributions switch.
\end{proof}

\section{Sound Explanation}

A sound explanation of a machine learning system should accurately and reliably
describe how the system makes a prediction. This work considers the problem of
explaining the system's prediction given a single input. A central part of our
definition is to include all the critical information used by the system as a
sound explanation. The information presentation should be in a
human-understandable way, but we do not attempt to define or quantify what is
understandable by humans. After achieving soundness, one can trade off between
explainability (i.e., how easy the explanation is to understand, which typically
relates to the complexity of the explanation) and prediction accuracy.

For example, consider the task of pancreatic cancer risk prediction given a
person's health records. Assume a two-step system with two black-box models $A$
and $B$. Model $A$ takes the raw health records of a patient and outputs values
indicating the presence or absence of a few meaningful concepts, such as
abdominal pain, personal history of cancer, and abnormal lab results. Model $B$
calculates the risk score only using the output of $A$. The output of $A$ is a
sound explanation of the system because one can verify that the system is not
using any information other than the presented explanation to make the
prediction.

This research does not consider the faithfulness of a sound explanation. A sound
explanation may fail to match the claimed semantics. In the above example, the
system can claim that a person has a higher pancreatic cancer risk due to recent
onset of abdominal pain. However, the presence of abdominal pain is a prediction
by another model, which may be incorrect from the original medical records.

Formally, we represent a machine learning system $\mlsys$ as a computational
graph $\mlsys = (G,\, P)$, where $G = (V,\, E)$ is a directed acyclic graph and
$P$ is a map from vertices to computable functions. Decompose $V = I \cup \{t\}
\cup U$, where $I$ represents the input variables, $t$ the output variable, and
$U$ the set of internal computation steps. Each non-input vertex $v \in V
\setminus I$ has an associated computable function $P_v$ defining its
functionality. (Only) Vertices in $I$ have no incoming edges, and (only) $t$ has
no outgoing edges. Let $x$ be an input to the system, which is a map from a
variable $v \in I$ to its value, denoted as $x_v$. Define a value function
$f_x(\cdot)$ for the vertices. For $v \in I$, $f_x(v) =x_v$. For other vertices
$v \in V \setminus I$, $f_x(v)$ is recursively computed as $f_x(v) = P_v(f_x(i):
(i,\, v) \in E)$ where $P_v$ is the computable function associated with $v$. The
output of $\mlsys$ on input $x$ is $f_x(t)$.

\begin{definition}[Sound explanation]
    \label{def:sound-expl}
    For a machine learning system $\mlsys = (G = (V = I \cup \{t\} \cup U,\, E),
    \,P)$, a sound explanation is a cut $C = (S,\,T)$ such that $I \subseteq S$,
    $t \in T$, $S \cap T = \emptyset$, and $S \cup T = V$. \\ Define $V_C \defeq
    \condSet{v \in S} {\exists v' \in T: (v,\,v')\in E}$ as the set of vertices
    at the boundary of the cut. Given an input $x$, define
    \begin{align*}
        \se_x \defeq \condSet{(v,\,f_x(v))}{v \in V_C}
    \end{align*}
    The information in $\se_x$ is presented to a human as the explanation of
    $f_x(t)$ on the input $x$.
\end{definition}

The design space of sound explanations is how to decompose a machine learning
system into a computational graph and a cut so that the explanation $\se$ is
meaningful to humans.

Our formulation incorporates inherently interpretable models. For example, for a
decision tree, a typical sound explanation is to present all the branching
decisions of the tree. In our framework, this can be implemented by converting
the tree into a computational graph where each tree node corresponds to a graph
vertex. Add edges to connect input variables to internal tree nodes. Add an
output node $t$ with incoming edges from all the leaf nodes. For a non-input
node~$v$, $f_x(v) \in \{0,\,1\}$ indicates whether this node is activated by the
input $x$. The sound explanation is the cut $C = (V \setminus \{t\},\, \{t\})$.
The explanation $\se_x$ shows which leaf node is activated for $x$,
corresponding to a path on the tree.

\section{An Example: Cancer Prediction}

We demonstrate sound explanation of models for cancer prediction. We work with
two cancer types: Pancreatic Ductal Adenocarcinoma (PDAC) and Hepatocellular
Carcinoma (HCC). Our datasets have thousands of features per patient with
millions of patients. Neural networks trained on the full dataset perform well,
but clinicians are reluctant to trust those models because even a linear model
with thousands of inputs is difficult to understand. We designed a scalable and
effective feature selection algorithm that resulted in \PDACreleaseNumFtr{}
features for PDAC and \HCCreleaseNumFtr{} features for HCC without a significant
performance drop. The selected features, which can be formulated as a binary
mask on the input in a computational graph per \cref{def:sound-expl}, serve as a
sound explanation. Since the lists of selected features correspond well to known
PDAC and HCC risk factors, clinicians gain higher confidence in our models.

\subsection{Task and Datasets}

We formulate cancer prediction as a binary classification task. Given the
Electronic Health Records (EHRs) of a patient and a cutoff date, the model
predicts risk of cancer diagnosis 6 to 18 months after the cutoff date from EHR
entries up to the cutoff date.

We derived our datasets from EHRs obtained on a federated network platform. The
platform aggregates EHRs from nearly 60 Healthcare Organizations (HCOs) in the
United States. EHRs have date-indexed entries of diagnosis, medication, and lab
records as well as demographic information including sex, birth year, and race
for each patient. To protect privacy, patients are anonymized, and dates are
perturbed. Data can only be accessed and processed within a secure computer
network.

For each type of cancer, we queried the platform for patients with certain
International Classification of Diseases (ICD) diagnosis codes to construct the
positive set and sampled patients without those ICD codes to construct the
negative set. For PDAC we used ICD-10 codes C25.0, C25.1, C25.2, C25.3, C25.7,
C25.8, and C25.9 and ICD-9 code 157. For HCC we used ICD-10 code C22.0 and ICD-9
code 155.0. We filtered the EHR data to improve data quality; for example, we
removed patients with entries two months after their recorded death date. To
compute the features, we filtered diagnosis, medication, and lab codes by
keeping those that appear in at least 1\% of the patients in the positive set.
We derived a fixed-length feature vector for each patient by calculating
features such as first date, last date, frequency, time span, medication
frequency, and lab value for each code. Our features also include age, sex, and
frequency of clinical encounters. All features except frequency of clinical
encounters have a corresponding binary existence feature to account for missing
data. Our PDAC dataset has \PDACtotPos{} positive cases and \PDACtotNeg{}
negative cases with \PDACftrDecayZeroNumFtr{} features. Our HCC dataset has
\HCCtotPos{} positive cases and \HCCtotNeg{} negative cases with
\HCCftrDecayZeroNumFtr{} features. The features are sparse. On average, each
feature has a 94\% chance of being zero.

\subsection{Method}

For each positive patient, we sampled a cutoff date uniformly 6 to 18 months
before their diagnosis date. For negative patients, we sampled cutoff dates
matched with the distribution of cutoff dates of positive patients. We derived
fixed-length features for each patient based on EHR entries up to the sampled
cutoff date. The cutoff dates were resampled for each minibatch during training.

We trained MLP neural networks with two hidden layers, each having 64 and 20
neurons with tanh nonlinearity, respectively. We applied $L_0$ regularization to
the weights to improve model generalizability. We extended the BinMask method
\citep{ jia2023effective} to select features: we applied $L_0$ regularization to
a parameterized binary input mask and obtained a feature mask by comparing the
smoothed mask with a threshold of 0.5. We then iteratively removed features
selected by BinMask. At each iteration, we evaluated the AUCs on the training
set with each feature removed and removed the feature that resulted in the
lowest AUC drop. This process was repeated until the AUC was 0.006 lower than
the AUC obtained before feature removal. We retrained the networks from random
initialization using the selected features.

\subsection{Results}

On the PDAC dataset, the model with all \PDACftrDecayZeroNumFtr{} features
achieved \PDACftrDecayZeroNNAuc{} AUC. BinMask selected \PDACftrDecaySelNumFtr{}
features. Iterative feature removal kept \PDACreleaseNumFtr{} features. The
final model achieved \PDACnnTestAucSingle{} AUC. Features include known PDAC
risk factors such as age, sex, diabetes mellitus, pancreatitis, pancreatic cysts,
abdominal pain, and smoking \citep{principe2020updated}, as well as novel
features associated with hypertension, hypercholesterolemia, kidney function,
and frequency of clinical visits.

On the HCC dataset, the model with all \HCCftrDecayZeroNumFtr{} features
achieved \HCCftrDecayZeroNNAuc{} AUC. BinMask selected \HCCftrDecaySelNumFtr{}
features. Iterative feature removal kept \HCCreleaseNumFtr{} features. The final
model achieved \HCCnnTestAucSingle{} AUC. Features include known HCC risk
factors such as cirrhosis, hepatitis, smoking, alcohol, age, and sex
\citep{yang2019global}. The model also utilizes features associated with
Nonalcoholic Fatty Liver Disease (NAFLD), the most common cause of HCC
developing in a non-cirrhotic liver \citep{gawrieh2019characteristics}.
NAFLD-related features include obesity, diabetes mellitus, hypertension, and
dyslipidemia, which correlate with the metabolic syndrome \citep{
kim2018nonalcoholic}. The features suggest that the model may predict HCC cases
that develop in a NAFLD liver, without underlying cirrhosis.

\cref{sec:ftr-list} presents the complete lists of selected features.

\bibliography{refs}

\begin{thebibliography}{20}
\providecommand{\natexlab}[1]{#1}
\providecommand{\url}[1]{\texttt{#1}}
\expandafter\ifx\csname urlstyle\endcsname\relax
  \providecommand{\doi}[1]{doi: #1}\else
  \providecommand{\doi}{doi: \begingroup \urlstyle{rm}\Url}\fi

\bibitem[Andrews et~al.(1995)Andrews, Diederich, and Tickle]{andrews1995survey}
Robert Andrews, Joachim Diederich, and Alan~B Tickle.
\newblock Survey and critique of techniques for extracting rules from trained
  artificial neural networks.
\newblock \emph{Knowledge-based systems}, 8\penalty0 (6):\penalty0 373--389,
  1995.

\bibitem[Broniatowski et~al.(2021)]{broniatowski2021psychological}
David~A Broniatowski et~al.
\newblock Psychological foundations of explainability and interpretability in
  artificial intelligence.
\newblock \emph{NIST, Tech. Rep}, 2021.

\bibitem[Doshi-Velez \& Kim(2017)Doshi-Velez and Kim]{doshi2017towards}
Finale Doshi-Velez and Been Kim.
\newblock Towards a rigorous science of interpretable machine learning.
\newblock \emph{arXiv preprint arXiv:1702.08608}, 2017.

\bibitem[Friedman \& Moulin(1999)Friedman and Moulin]{friedman1999three}
Eric Friedman and Herve Moulin.
\newblock Three methods to share joint costs or surplus.
\newblock \emph{Journal of economic Theory}, 87\penalty0 (2):\penalty0
  275--312, 1999.

\bibitem[Friedman(2004)]{friedman2004paths}
Eric~J Friedman.
\newblock Paths and consistency in additive cost sharing.
\newblock \emph{International Journal of Game Theory}, 32:\penalty0 501--518,
  2004.

\bibitem[Gawrieh et~al.(2019)Gawrieh, Dakhoul, Miller, Scanga, DeLemos,
  Kettler, Burney, Liu, Abu-Sbeih, Chalasani,
  et~al.]{gawrieh2019characteristics}
Samer Gawrieh, Lara Dakhoul, Ethan Miller, Andrew Scanga, Andrew DeLemos, Carla
  Kettler, Heather Burney, Hao Liu, Hamzah Abu-Sbeih, Naga Chalasani, et~al.
\newblock Characteristics, aetiologies and trends of hepatocellular carcinoma
  in patients without cirrhosis: a united states multicentre study.
\newblock \emph{Alimentary pharmacology \& therapeutics}, 50\penalty0
  (7):\penalty0 809--821, 2019.

\bibitem[Jia \& Rinard(2023)Jia and Rinard]{jia2023effective}
Kai Jia and Martin Rinard.
\newblock Effective neural network $l_0$ regularization with binmask.
\newblock \emph{arXiv preprint arXiv:2304.11237}, 2023.

\bibitem[Kim et~al.(2018)Kim, Touros, and Kim]{kim2018nonalcoholic}
Donghee Kim, Alexis Touros, and W~Ray Kim.
\newblock Nonalcoholic fatty liver disease and metabolic syndrome.
\newblock \emph{Clinics in liver disease}, 22\penalty0 (1):\penalty0 133--140,
  2018.

\bibitem[Koh et~al.(2020)Koh, Nguyen, Tang, Mussmann, Pierson, Kim, and
  Liang]{koh2020concept}
Pang~Wei Koh, Thao Nguyen, Yew~Siang Tang, Stephen Mussmann, Emma Pierson, Been
  Kim, and Percy Liang.
\newblock Concept bottleneck models.
\newblock In \emph{International Conference on Machine Learning}, pp.\
  5338--5348. PMLR, 2020.

\bibitem[Krishna et~al.(2022)Krishna, Han, Gu, Pombra, Jabbari, Wu, and
  Lakkaraju]{krishna2022disagreement}
Satyapriya Krishna, Tessa Han, Alex Gu, Javin Pombra, Shahin Jabbari, Steven
  Wu, and Himabindu Lakkaraju.
\newblock The disagreement problem in explainable machine learning: A
  practitioner's perspective.
\newblock \emph{arXiv preprint arXiv:2202.01602}, 2022.

\bibitem[Kumar et~al.(2022)Kumar, Yu, Kannampallil, Abrams, Michelson, and
  Payne]{kumar2022self}
Sayantan Kumar, Sean~C Yu, Thomas Kannampallil, Zachary Abrams, Andrew
  Michelson, and Philip~RO Payne.
\newblock Self-explaining neural network with concept-based explanations for
  icu mortality prediction.
\newblock In \emph{Proceedings of the 13th ACM International Conference on
  Bioinformatics, Computational Biology and Health Informatics}, pp.\  1--9,
  2022.

\bibitem[Molnar(2020)]{molnar2020interpretable}
Christoph Molnar.
\newblock \emph{Interpretable machine learning}.
\newblock Independently published, 2020.

\bibitem[Park et~al.(2018)Park, Hendricks, Akata, Rohrbach, Schiele, Darrell,
  and Rohrbach]{park2018multimodal}
Dong~Huk Park, Lisa~Anne Hendricks, Zeynep Akata, Anna Rohrbach, Bernt Schiele,
  Trevor Darrell, and Marcus Rohrbach.
\newblock Multimodal explanations: Justifying decisions and pointing to the
  evidence.
\newblock In \emph{Proceedings of the IEEE conference on computer vision and
  pattern recognition}, pp.\  8779--8788, 2018.

\bibitem[Principe \& Rana(2020)Principe and Rana]{principe2020updated}
Daniel~R Principe and Ajay Rana.
\newblock Updated risk factors to inform early pancreatic cancer screening and
  identify high risk patients.
\newblock \emph{Cancer letters}, 485:\penalty0 56--65, 2020.

\bibitem[Ribeiro et~al.(2016)Ribeiro, Singh, and Guestrin]{ribeiro2016should}
Marco~Tulio Ribeiro, Sameer Singh, and Carlos Guestrin.
\newblock "why should i trust you?" explaining the predictions of any
  classifier.
\newblock In \emph{Proceedings of the 22nd ACM SIGKDD international conference
  on knowledge discovery and data mining}, pp.\  1135--1144, 2016.

\bibitem[Rudin(2019)]{rudin2019stop}
Cynthia Rudin.
\newblock Stop explaining black box machine learning models for high stakes
  decisions and use interpretable models instead.
\newblock \emph{Nature machine intelligence}, 1\penalty0 (5):\penalty0
  206--215, 2019.

\bibitem[Sudjianto \& Zhang(2021)Sudjianto and Zhang]{sudjianto2021designing}
Agus Sudjianto and Aijun Zhang.
\newblock Designing inherently interpretable machine learning models.
\newblock \emph{arXiv preprint arXiv:2111.01743}, 2021.

\bibitem[Sundararajan et~al.(2017)Sundararajan, Taly, and
  Yan]{sundararajan2017axiomatic}
Mukund Sundararajan, Ankur Taly, and Qiqi Yan.
\newblock Axiomatic attribution for deep networks.
\newblock In \emph{International conference on machine learning}, pp.\
  3319--3328. PMLR, 2017.

\bibitem[Yang et~al.(2019)Yang, Hainaut, Gores, Amadou, Plymoth, and
  Roberts]{yang2019global}
Ju~Dong Yang, Pierre Hainaut, Gregory~J Gores, Amina Amadou, Amelie Plymoth,
  and Lewis~R Roberts.
\newblock A global view of hepatocellular carcinoma: trends, risk, prevention
  and management.
\newblock \emph{Nature reviews Gastroenterology \& hepatology}, 16\penalty0
  (10):\penalty0 589--604, 2019.

\bibitem[Zeiler \& Fergus(2014)Zeiler and Fergus]{zeiler2014visualizing}
Matthew~D Zeiler and Rob Fergus.
\newblock Visualizing and understanding convolutional networks.
\newblock In \emph{Computer Vision--ECCV 2014: 13th European Conference,
  Zurich, Switzerland, September 6-12, 2014, Proceedings, Part I 13}, pp.\
  818--833. Springer, 2014.

\end{thebibliography}
\bibliographystyle{tmlr}

\newpage
\appendix
\onecolumn

\section{Complete lists of selected features for cancer prediction
\label{sec:ftr-list}}

\begin{figure*}[t]
    \centering
    \includegraphics[width=.9\textwidth]{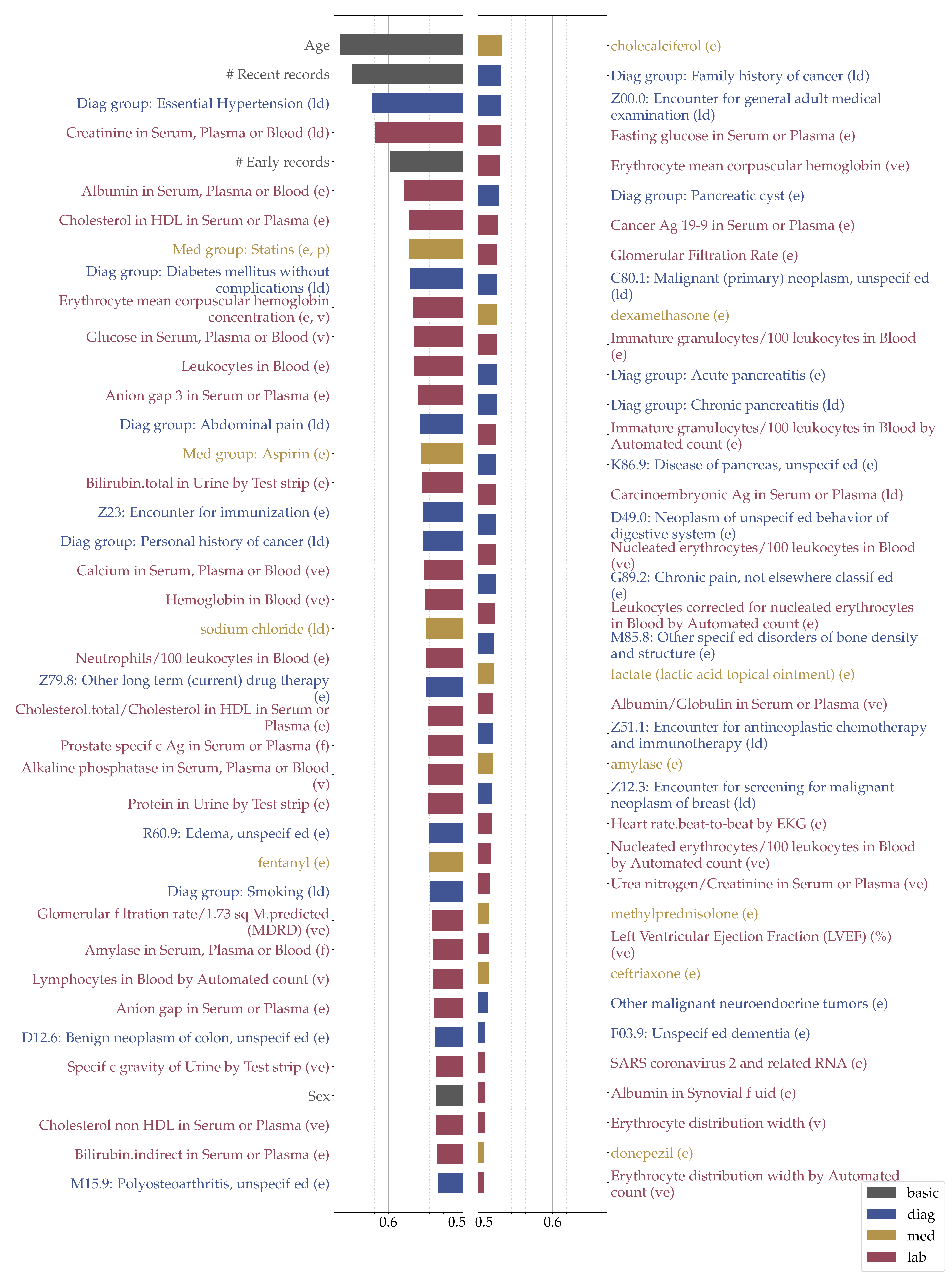}
    \caption{List of selected features ranked by univariate model AUC for PDAC
        \label{fig:PDACftr}
    }
\end{figure*}

\begin{figure*}[t]
    \centering
    \includegraphics[width=.9\textwidth]{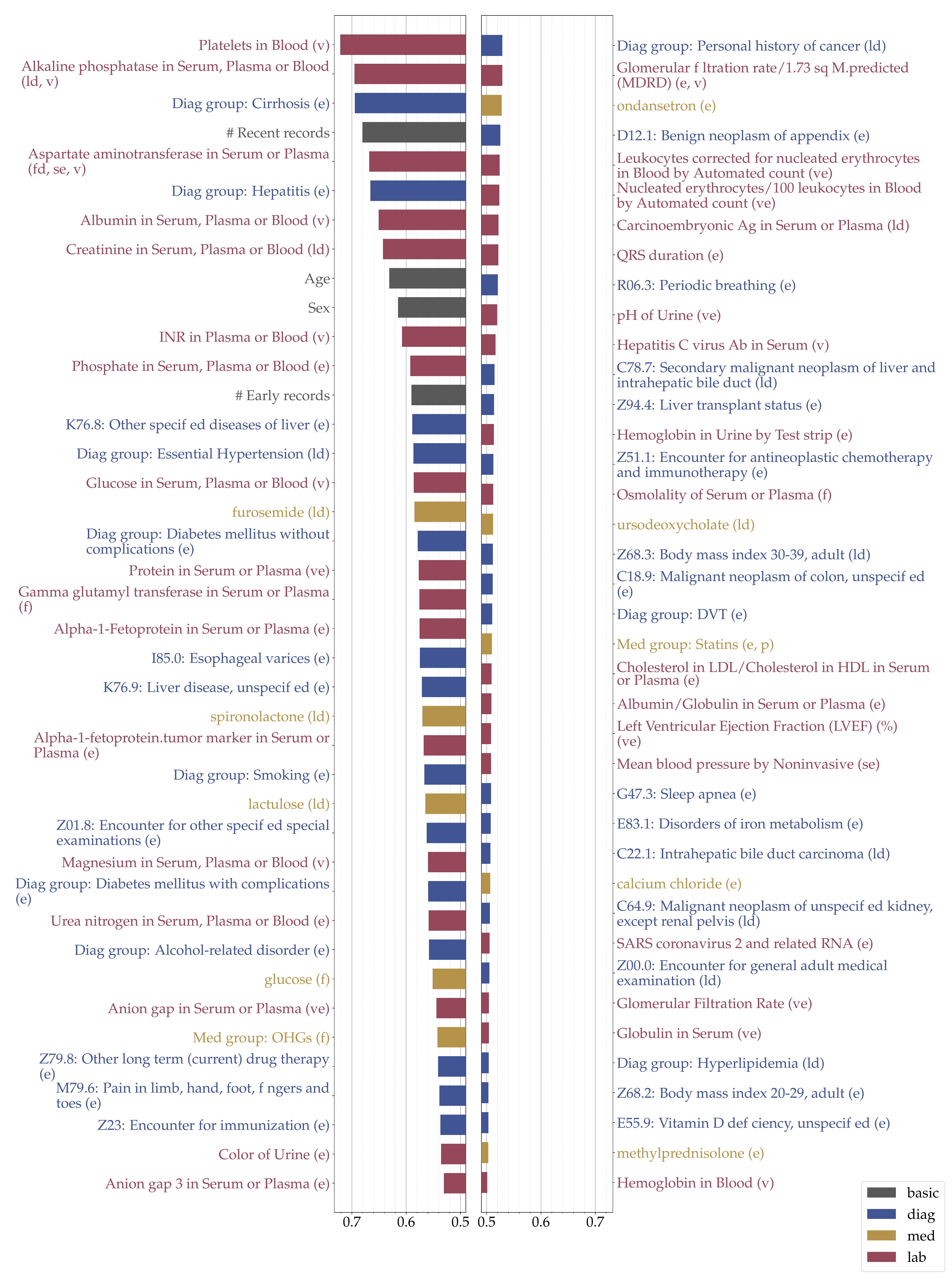}
    \caption{List of selected features ranked by univariate model AUC for HCC
        \label{fig:HCCftr}
    }
\end{figure*}

This section presents the complete lists of selected features. We define a
heuristic score, the \emph{univariate model AUC}, to rank the features. The
univariate model AUC represents the predictive power of a single feature on the
test set. It is computed by evaluating the model AUC on the test set with values
of each feature while setting all other features to zero. \cref{fig:PDACftr}
presents the list of PDAC features, and \cref{fig:HCCftr} presents the list of
HCC features. In the plots, the label \texttt{diag} refers to diagnosis features,
\texttt{med} to medication features, and \texttt{lab} to lab features. Letters
in the brackets indicate the types of derived features: \texttt{e} for existence,
\texttt{fd} for first date, \texttt{ld} for last date, \texttt{p} for time span,
\texttt{f} for frequency, \texttt{v} for latest lab value, \texttt{ve} for
whether a valid lab value exists, \texttt{s} for lab value slope, and \texttt{se}
for whether lab value slope can be computed.

\end{document}